\documentclass[sn-mathphys]{sn-jnl}% Math and Physical Sciences Reference Style
\usepackage{comment}
\usepackage{epstopdf}
\usepackage{subfigure}
\usepackage{indentfirst}

\usepackage{textcomp,booktabs}
\usepackage{pifont}% http://ctan.org/pkg/pifont
\newcommand{\cmark}{\ding{51}}%
\newcommand{\xmark}{\ding{55}}%

\jyear{2023}%

\theoremstyle{thmstyleone}%
\theoremstyle{thmstyletwo}%

\theoremstyle{thmstylethree}%

\begin{document}

\title[Human Activity Recognition using RGB-Event based Sensors]{Human Activity Recognition using RGB-Event based Sensors: A Multi-modal Heat Conduction Model and A Benchmark Dataset} 

% HARDVS: Revisiting Human Activity Recognition with Dynamic Vision Sensors 

\author[1]{\fnm{Shiao} \sur{Wang}}\email{e24101001@stu.ahu.edu.cn}
\author*[1]{\fnm{Xiao} \sur{Wang}}\email{xiaowang@ahu.edu.cn}
\author*[1]{\fnm{Bo} \sur{Jiang}}\email{jiangbo@ahu.edu.cn}
%\equalcont{These authors contributed equally to this work.}
\author[2]{\fnm{Lin} \sur{Zhu}} \email{linzhu@pku.edu.cn}
\author[3]{\fnm{Guoqi} \sur{Li}} \email{guoqi.li@ia.ac.cn}
\author[4,5]{\fnm{Yaowei} \sur{Wang}}\email{wangyw@pcl.ac.cn}
\author[5,6,7]{\fnm{Yonghong} \sur{Tian}}\email{yhtian@pku.edu.cn}
\author[1]{\fnm{Jin} \sur{Tang}}\email{tangjin@ahu.edu.cn}

\affil[1]{\orgdiv{School of Computer Science and Technology, Anhui University, Hefei, 230601, China}}
\affil[2]{\orgdiv{Beijing Institute of Technology, Beijing, China}}
\affil[3]{\orgdiv{University of Chinese Academy of Sciences, Beijing, China}}
\affil[4]{\orgdiv{Harbin Institute of Technology, Shenzhen, China}}
\affil[5]{\orgdiv{Peng Cheng Laboratory, Beijing, China}}
\affil[6]{\orgdiv{National Key Laboratory for Multimedia Information Processing, School of Computer Science, Peking University, China}}
\affil[7]{\orgdiv{School of Electronic and Computer Engineering, Shenzhen Graduate School, Peking University, China}}

\abstract{
Human Activity Recognition (HAR) has long been a fundamental research direction in the field of computer vision. Previous studies have primarily relied on traditional RGB cameras to achieve high-performance activity recognition. However, the challenging factors in real-world scenarios, such as insufficient lighting and rapid movements, inevitably degrade the performance of RGB cameras. 
To address these challenges, biologically inspired event cameras offer a promising solution to overcome the limitations of traditional RGB cameras. In this work, we rethink human activity recognition by combining the RGB and event cameras. The first contribution is the proposed large-scale multi-modal RGB-Event human activity recognition benchmark dataset, termed HARDVS 2.0, which bridges the dataset gaps. It contains 300 categories of everyday real-world actions with a total of 107,646 paired videos covering various challenging scenarios. 
Inspired by the physics-informed heat conduction model, we propose a novel multi-modal heat conduction operation framework for effective activity recognition, termed MMHCO-HAR. More in detail, given the RGB frames and event streams, we first extract the feature embeddings using a stem network. Then, multi-modal Heat Conduction blocks are designed to fuse the dual features, the key module of which is the multi-modal Heat Conduction Operation (HCO) layer. We integrate RGB and event embeddings through a multi-modal DCT-IDCT layer while adaptively incorporating the thermal conductivity coefficient via FVEs (Frequency Value Embeddings) into this module. After that, we propose an adaptive fusion module based on a policy routing strategy for high-performance classification. 
We conduct comprehensive experiments comparing our proposed method with baseline methods on the HARDVS 2.0 dataset and other public datasets. These results demonstrate that our method consistently performs well, validating its effectiveness and robustness. 
The source code and benchmark dataset will be released on~\url{https://github.com/Event-AHU/HARDVS/tree/HARDVSv2}. 
}


\maketitle

\section{Introduction}~\label{sec1}
Human-centered visual tasks (e.g., human activity recognition~\cite{chen2021deepHARsurvey}, pedestrian attribute recognition~\cite{wang2022PARsurvey}, person re-identification~\cite{sarker2024transformerReIDSurvey}) have increasingly garnered attention with the development of deep learning and computer vision. 
For the critical task of human activity recognition~\cite{kong2018humanARSurvey, ahmad2021graph, zhang2019comprehensive, ji20123d, yao2011human, varol2021synthetic}, most researchers rely on the mature technology of RGB cameras to achieve effective activity classification. 
In recent years, numerous activity recognition tasks based on RGB cameras have emerged, involving various application scenarios such as intelligent surveillance, sports activity analysis, and human-computer interaction. However, the inherent limitations of the traditional RGB sensors pose various challenges in practical application, including issues related to data usage, analysis, and ethics. 
On the one hand, the standard RGB cameras often have limited frame rates (e.g., 30FPS), resulting in motion blur when capturing fast-moving scenes, such as athletes performing high-speed activities in sports events. 
On the other hand, when facing extreme light conditions such as overexposure or low-light, traditional RGB cameras' low dynamic range tends to produce low-quality videos. 
% Furthermore, the usage and transmission of the datasets collected by RGB cameras raise ethical concerns related to privacy protection. 
These challenges significantly hinder the progress of current human activity recognition tasks. 
% Naturally, this prompts the question: \emph{can we leverage novel sensors to address the shortcomings of RGB cameras?}

Recently, bio-inspired sensors (also termed event-based cameras), 
such as DAVIS~\cite{brandli2014240}, CeleX-V~\cite{chen2019celexV}, ATIS~\cite{posch2010qvga}, 
and PROPHESEE~\footnote{\url{https://www.prophesee.ai}}, drawing more and more researchers' attention and have been introduced for research on pattern recognition, object detection and tracking~\cite{wang2024eventTrack, li2023semantic, yuan2023learning, wang2021visevent, lu2024modality, zhang2024learning}. 
Unlike the synchronous optical imaging principle of RGB cameras, event cameras record event signals generated by changes in illumination caused by the movement of objects in an asynchronous manner. Specifically, each pixel in the field of view of the event cameras independently records a binary signal generated by lighting increases or decreases (if and only if a certain threshold is exceeded), which is commonly referred to as polarity. Due to the unique imaging principle of event cameras, they typically offer sparse spatial resolution and dense temporal resolution. As a result, event cameras tend to provide stable imaging when capturing rapidly changing human movements, such as the fast hand movements of magicians and the swift sports activities of an athlete, without causing motion blur. Additionally, since event signals are generated by changes in illumination, the event cameras do not require strict lighting conditions, exhibiting a high dynamic range (HDR) advantage. Thus, they can perform well in overexposed and low-light scenarios. 
% Furthermore, the event streams signals produced by event cameras only contain the contour information of the target's motion, which can significantly protect individual privacy in human-centered tasks, thereby avoiding ethical issues arising from privacy leakage. 
In light of the aforementioned issues with RGB cameras and the benefits of the unique advantages of event cameras, we consider combining event cameras with RGB cameras to address the challenges in human activity recognition tasks.

\begin{figure} 
\center
\includegraphics[width=4.6in]{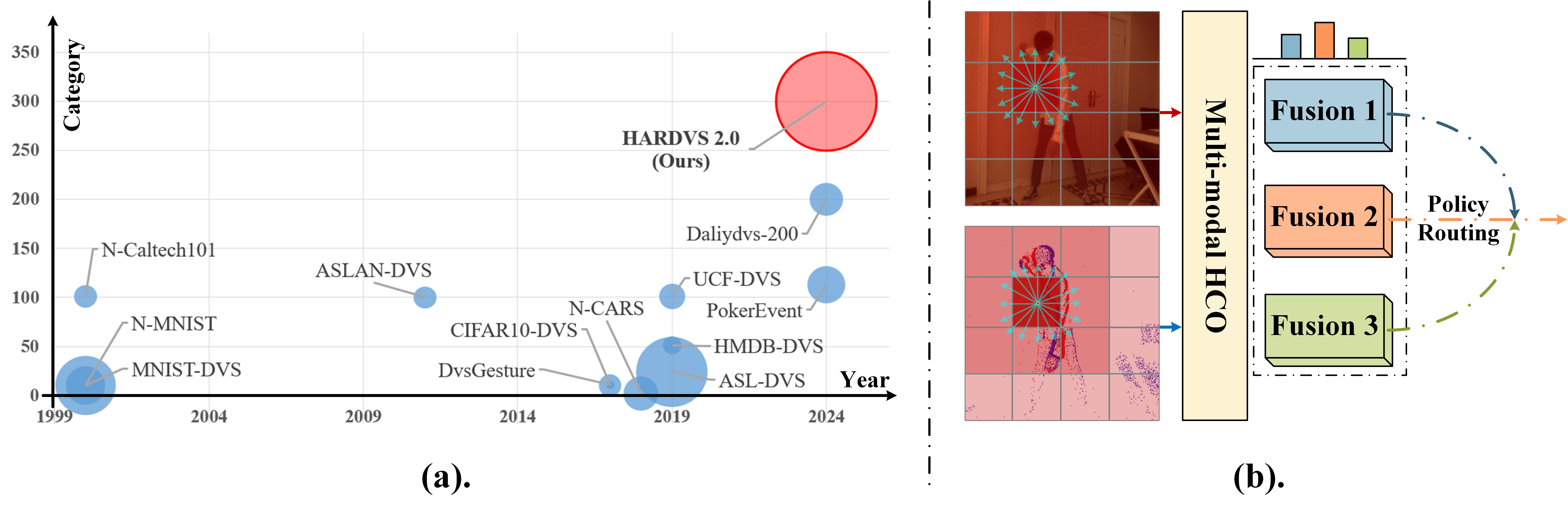}
\caption{(a). Comparison between existing datasets and our proposed HARDVS 2.0 dataset for video classification. (b). A simple schematic diagram of our framework.} 
\label{benchmarkCOMP}
\end{figure}

Since this research direction is still in its early stages, there are not many open-source public datasets available in academia. Although several benchmark datasets have been proposed for classification tasks~\cite{bi2020graph, amir2017low, li2017cifar10, serrano2015poker, kuehne2011hmdb, soomro2012ucf101, kliper2011action, planamente2021da4event, cannici2021nrod, xiao2024cs, liu2025comprehensive}, most of them are simulated or synthetic datasets derived from RGB videos using simulators. 
Some researchers also obtain event data by recording screens while displaying RGB videos. However, these datasets fail to capture the real-world characteristics of event cameras, particularly in fast-motion and low-light scenarios. 
The ASL-DVS dataset~\cite{bi2020graph} proposed by Bi et al. consists of $100,800$ samples but is limited to hand gesture recognition with only 24 classes. DvsGesture~\cite{amir2017low} is also constrained in scale and category coverage, which limits its relevance in the context of deep learning. 
Furthermore, some datasets, like DvsGesture, have reached performance saturation; for instance, Wang et al.~\cite{wang2019steventclouds} achieved a recognition accuracy of $97.08\%$ on DvsGesture~\cite{amir2017low}. 
As a result, there remains a strong demand in the research community for a large-scale, real-world human activity recognition benchmark dataset.

In this paper, we propose a large-scale RGB-Event benchmark dataset called HARDVS 2.0 to address the problem of the lack of real RGB-Event data for human activity recognition. Specifically, our proposed HARDVS 2.0 dataset includes over $100,000$ video clips, each lasting approximately 5 to 10 seconds, recorded with a DAVIS346 event camera. It covers 300 categories of human activities in daily life, such as drinking, {riding a bike}, {sitting down}, and {washing hands}. To ensure diversity, the dataset incorporates a variety of factors, including {multi-view} perspectives, varying {illumination conditions}, {motion speeds}, {dynamic backgrounds}, {occlusion}, {flashing light}, and {photographic distance}. To the best of our knowledge, HARDVS 2.0 is the first realistic, large-scale, and challenging benchmark multi-modal dataset for human activity recognition in real-world, uncontrolled environments. A comparison of existing recognition datasets with our HARDVS 2.0 is shown in Fig.~\ref{benchmarkCOMP} (a).

Based on our newly proposed multi-modal dataset HARDVS 2.0, we design a novel framework for \textbf{M}ulti-\textbf{M}odal visual \textbf{H}eat \textbf{C}onduction \textbf{O}peration based \textbf{HAR} task, termed \textbf{MMHCO-HAR}. 
Inspired by the physics-informed heat conduction and the success of the heat conduction operation-based vision model~\cite{wang2025object, wang2024vheat}, in this work, we propose a novel multi-modal heat conduction operation framework for efficient activity recognition, termed MMHCO-HAR. 
Specifically speaking, we first adopt a stem network to transform the input RGB frames and event streams into corresponding feature embeddings. The multi-modal heat conduction blocks are proposed to fuse the dual features, the key module of which is the multi-modal Heat Conduction Operation (HCO) layer. In our implementation, we adopt the multi-modal DCT-IDCT layer to integrate RGB and event embeddings and incorporate the heat conductivity coefficient via FVEs (Frequency Value Embeddings) into this module adaptively. Then, we design three different feature fusion strategies for various feature combinations in diverse situations and utilize a policy routing network to select a fusion strategy adaptively, aiming to alleviate the issue of imbalanced multi-modal learning. A simple schematic diagram of our framework is visualized in Fig.~\ref{benchmarkCOMP} (b).

To sum up, the main contributions of this paper can be summarized as the following three aspects: 

$\bullet$ We propose a large-scale benchmark dataset for RGB-Event based human activity recognition, termed HARDVS 2.0. To the best of our knowledge, it is the first large-scale multi-modal dataset for HAR, which contains 300 categories of everyday real-world actions with a total of 107,646 paired videos covering various challenging scenarios.

$\bullet$ We propose a multi-modal heat conduction-based backbone network for human activity recognition. Our RGB-Event HAR framework achieves lower computational complexity, higher performance, and better physical interpretability compared with existing models. 

$\bullet$ We establish a benchmark on the HARDVS 2.0 dataset, offering a robust platform for future works to compare. Extensive experiments conducted on HARDVS 2.0, along with other widely used benchmark datasets, comprehensively demonstrate the effectiveness of our proposed model.

A preliminary version of this work was published at the international conference, i.e., Association for the Advancement of Artificial Intelligence (AAAI) 2024. Compared with the conference version~\cite{wang2024hardvs}, we make the following extensions: 
\textbf{1). A Novel Large-Scale Multi-Modal HAR Dataset:} 
The previous conference version focused on human action recognition using pure event streams. In this paper, we extend this work to multi-modal scenarios and release the entire multi-modal dataset to further support research in this area.
\textbf{2). New Multi-modal Heat Conduction-based Visual Framework:} 
In the previous conference version, we proposed a Transformer-based approach for spatio-temporal modeling of event streams. In this paper, we introduce a novel multi-modal HAR backbone network inspired by physical heat conduction, which achieves higher recognition accuracy, lower computational complexity, and enhanced physical interpretability.
\textbf{3). A Policy Routing Based Fusion Method:} We propose a novel policy network and routing mechanism based fusion method, which alleviates the unbalanced multi-modal issue. 
\textbf{4). More Extensive Experiments:} We have conducted more comprehensive experiments to validate the effectiveness of our proposed method. Our approach has seen substantial improvements compared to its initial version.

The rest of this paper is organized as follows: 
In section~\ref{sec2}, we review the related works based on event based HAR, RGB-Event based HAR, Biology and Physics Inspired Models, and Benchmark Datasets for HAR. 
In section~\ref{sec3}, we introduce our proposed HAR framework using RGB-Event sequences based on physical heat conduction. 
In section~\ref{sec4}, we describe the collection protocols and statistical analysis of the HARDVS 2.0 dataset. 
After that, in section~\ref{sec5}, we conduct experiments to evaluate our proposed HAR framework from both quantitative and qualitative analysis perspectives. 
Finally, we conclude this paper and propose possible research directions as our future works in section~\ref{sec6}.

\section{Related Work}~\label{sec2}

In this section, we give a brief review of the event Camera-based Human Activity Recognition Methods, RGB-Event based HAR, Biology and Physics Inspired Models, and event benchmark datasets for HAR. More works about HAR and event cameras can be found in the following surveys~\cite{kong2018humanARSurvey, ahmad2021graph, gallegoevent}. 

\subsection{Event based HAR}
Human activity recognition tasks typically rely on traditional RGB cameras to achieve high-performance classification. However, the effect often decreases under unfavorable lighting conditions or when the movement is too fast. Many researchers have focused on using event cameras to improve the effectiveness of human activity recognition tasks. 
Arnon et al.~\cite{amir2017low} propose for the first time a low-power fully event-driven gesture recognition system based on event hardware, which uses a TrueNorth neural morphology processor to achieve end-to-end processing and real-time recognition of gestures in events streamed by dynamic visual sensors (DVS). 
Xavier et al.~\cite{clady2017motion} propose an event-driven neuromorphic retina output without brightness features, which maps the optical flow distribution of moving objects in the field of view to a matrix for event-based pattern recognition. 
Chen et al.~\cite{chen2021novel} propose a robust gesture recognition system based on event-driven neural morphological vision sensors and active LED-labeled gloves, which can maintain high recognition accuracy under different lighting and background conditions. 
Chen et al.~\cite{chen2019fast} propose an event-driven fast retinal morphology representation method (EDR) that achieves real-time inference and learning in video games and activity recognition. 
Xavier et al.~\cite{lagorce2016hots} represent the recent temporal activity within a local spatial neighborhood, and utilize the rich temporal information provided by events to create contexts in the form of time surfaces, termed HOTS, for the recognition task. 
Wu et al.~\cite{wu2020multipath} first transform the event streams into images, then predict and combine the human pose with event images for HAR.
Graph neural networks (GNN) and SNNs are also exploited for event-based recognition. Specifically, 
Chen et al.~\cite{chen2020dyGCN} treat the event streams as a 3D point cloud and use dynamic GNNs to learn the spatial-temporal features for gesture recognition. 
Wang et al.~\cite{wang2021eventGNN} investigate the event streams representation method for recognizing human gait using deep neural networks. Two different event streams representations were proposed: image-based representation and graph based representation, and gait recognition was performed using graph convolutional networks and convolutional neural networks, respectively.
Xing et al.~\cite{xing2020new} propose a novel event based spiking convolutional recurrent neural network (SCRNN) that utilizes convolution operations and recurrent connectivity to process asynchronous and sparse event sequence data, effectively improving the accuracy of event-based gesture recognition. 
Different from these methods, we are considering combining event streams with RGB modality to achieve more accurate activity recognition using bimodal RGB-E data and model.

\subsection{RGB-Event based HAR}
To address the limited performance of traditional RGB cameras in extreme environments and enhance the representation capability of event data, the RGB-E based multi-modal human activity recognition task has been extensively explored by researchers.
Huang et al.~\cite{huang2022vefnet} propose VEFNet, a cross-modal fusion network that combines event streams and RGB images for visual position recognition, effectively addressing the challenges posed by lighting and seasonal changes, and achieving long-term localization.
Wang et al.~\cite{wang2023sstformer} integrate RGB frames and event streams by using a memory supported transformer network and a pulse neural network, a multi-modal bottleneck fusion module for feature aggregation, the current problems in event camera pattern recognition are solved.
Li et al.~\cite{li2023semantic} propose a new pattern recognition framework that integrates semantic labels, RGB frames, and event streams, utilizing pre-trained large-scale visual language models to address the semantic gap and small-scale backbone network issues present in existing methods.
TSCFormer proposed by Wang et al.~\cite{wang2023unleashingpowercnntransformer}, which is a relatively lightweight CNN and Transformer model. By bridging the Transformer module and interactive feature fusion module, it achieves the capture of large-scale global relationships between RGB-E modalities while maintaining a simple model structure.
For all that, the above studies have been evaluated primarily on the powerful global modeling capability of the Transformer network. However, due to the quadratic complexity of the attention mechanism, especially when multi-modal data are input, it will impose a heavy computational burden on the network during both the training and testing phases.

\subsection{Biology and Physics Inspired Models}
Convolutional neural networks (CNNs) are widely adopted and applied across various visual tasks~\cite{kayalibay2017cnn, ciocca2018cnn, mahmoudi2019multi}. To overcome the inherent limitation of CNNs, which have a local receptive field, a milestone visual Transformer~\cite{dosovitskiy2020image} network was developed to create a global receptive field for images. However, Transformers are often constrained by the quadratic complexity of their attention mechanisms. As a result, visual Mamba models~\cite{zhu2024vision, liu2024vmambavisualstatespace}, which are based on state space models and offer linear complexity, have gained popularity for a range of visual tasks. Despite their efficiency, the parallelization of the selective scanning mechanism in Mamba lacks interpretability on GPUs and fails to deliver optimal performance. In addition to the various popular neural network models mentioned above, biology and physics-inspired visual network models have also been widely explored. Spiking Neural Networks(SNNs)~\cite{ghosh2009spiking, tavanaei2019deep} aim to bridge the gap between neuroscience and machine learning by using models that best fit the mechanisms of biological neurons for computation, closer to the mechanisms of biological neurons for simple visual applications~\cite{yamazaki2022spiking}. Diffusion Models~\cite{ho2022video, xing2024survey} inspired by non-equilibrium thermodynamics, the theory first defines the Markov chain of diffusion steps to slowly add random noise to the data and then learns the reverse diffusion process. QB-Heat~\cite{chen2022self} utilizes the heat conduction equation for self-supervised learning, especially in the field of image feature learning. vHeat~\cite{wang2024vheat} treats image patches as heat sources and computes their correlation as the diffusion of thermal energy. The high computational efficiency and the global receptive field are achieved. In this paper, we design a novel multi-modal human activity recognition method based on the heat conduction model with physical principles, which has high interpretability and lower complexity.

\subsection{Benchmark Datasets for HAR~} 
As listed in Table~\ref{datasetlist}, we compared existing benchmark datasets based on event cameras for human activity recognition tasks. The early datasets listed in the table are either limited in the number of samples or synthetic datasets, which makes it difficult to reflect the characteristics of event cameras. For example, the N-Caltech101~\cite{orchard2015converting} and N-MNIST~\cite{orchard2015converting} datasets were recorded using an ATIS camera and contain 101 and 10 classes, respectively. Additionally, Bi et al.~\cite{bi2020graph} transformed popular HAR datasets into simulated event streams, including HMDB-DVS~\cite{bi2020graph, kuehne2011hmdb}, UCF-DVS~\cite{bi2020graph, soomro2012ucf101}, and ASLAN-DVS~\cite{kliper2011action}, thereby expanding the available dataset pool for HAR. However, these simulated event datasets do not fully capture the advantages of event cameras, such as performance under low light conditions or during fast motion. There are also four real-world event datasets for classification: DvsGesture~\cite{amir2017low}, N-CARS~\cite{sironi2018hats}, ASL-DVS~\cite{bi2020graph}, and PAF~\cite{miao2019neuromorphic}, but they are constrained by factors such as limited scale, category diversity, and scene variation. Concretely, these datasets contain only 11, 2, 24, and 10 classes, respectively, and rarely account for challenging factors like multi-view perspectives, motion dynamics, or visual noise. Recently, some relatively large real-world HAR datasets have also emerged, i.e., the PokerEvent~\cite{wang2023sstformer} and Dailydvs-200~\cite{wang2024dailydvs}. These datasets not only include tens of thousands of video sequences but also feature 114 and 200 activity categories, respectively. Building on this trend, we introduce a larger (100K samples) and more diverse (300 classes) real-world RGB-Event camera based human activity recognition dataset, named HARDVS 2.0. Our proposed dataset comprehensively includes various human activities in real life through indoor/outdoor shooting, fully reflecting the multiple challenge attributes mentioned above. We believe that the HARDVS 2.0 dataset will further promote the development of event cameras in the field of HAR.

\section{Methodology}~\label{sec3}

In this section, we will introduce a novel Heat Conduction-based human activity recognition method, termed MMHCO-HAR. Firstly, we give a preliminary introduction of physical heat conduction and its visual adaptation vHeat model. 
Then, an overview of our designed RGB-Event human activity recognition framework is introduced. 
After that, we describe the input encoding of RGB and event streams utilized in this work. 
Then, we delve into the details of our network architecture and loss functions. More details of each module will be described in the following sub-sections, respectively.

\subsection{Preliminaries: A Physics-Inspired Heat Conduction Model} 
Heat conduction has always been a classic problem in physics, which usually occurs in solids, liquids, and gases, but the thermal diffusivity varies among different substances. The vHeat~\cite{wang2024vheat} model transfers the heat conduction effect to visual tasks. In a two-dimensional image area, using point $(x, y)$ as the heat source and the thermal diffusivity is $k$ (where $k>0$), at time $t$, its temperature can be expressed as $u (x, y, t)$. The general solution of the heat conduction equation can be expressed as,

\begin{equation}
\label{HC_formula}
    u(x,y,t)=\mathcal{F}^{-1}(\widetilde{f}(\omega_x,\omega_y)e^{-k(\omega_x^2+\omega_y^2)t}),
\end{equation}
where $\mathcal{F}^{-1}$ denote the inverse Fourier Transform, $\widetilde{f}(\omega_x,\omega_y)$ is the representation of the initial conditions in the frequency domain, and $(\omega_x,\omega_y)$ denote the corresponding spatial frequency variables. $e^{-k(\omega_x^2+\omega_y^2)t}$ is a decay factor that decreases with increasing time $t$, reflecting the diffusion of thermal energy over time. 
Furthermore, in the case of 2D visual images, the formula~\ref{HC_formula} can be written as,
\begin{equation}
\label{visual_hc}
    U^t=\mathcal{F}^{-1}(\mathcal{F}(U^0)e^{-k(\omega_x^2+\omega_y^2)t}),
\end{equation}
where $U^0$ and $U^t$ denote the input features of given image patches and the output features. 
Since image patches exist in discrete form and are rectangles constrained by boundary conditions, two-dimensional discrete cosine transform (DCT) and two-dimensional inverse discrete cosine transform (IDCT) can be used instead of the Fourier transform and inverse discrete Fourier transform. Finally, the solution of the heat conduction equation in visual 2D images can be expressed as,
\begin{equation}
\label{DCT_IDCT}
    U^t=\mathbf{IDCT_{2D}}(\mathbf{DCT_{2D}}(U^0)e^{-k(\omega_x^2+\omega_y^2)t}).
\end{equation}
The above preliminaries provide a brief overview of heat conduction from a physical to a visual perspective. For a more detailed explanation of the background of the dynamic heat conduction, please refer to Non-equilibrium thermodynamics~\cite{de2013non} and vHeat~\cite{wang2024vheat}.

\subsection{Overview}
As shown in Fig.~\ref{framework}, we propose a novel heat conduction-based multi-modal learning framework for efficient and effective RGB-Event based human activity recognition. Concretely, we first adopt a stem network to transform the input RGB frames and event streams into corresponding feature embeddings. Then, the multi-modal HCO blocks are proposed to achieve RGB and event feature learning and interaction simultaneously. The core operation is the DCT-IDCT transformation network equipped with modality-specific continuous Frequency Value Embeddings (FVEs). After that, we explore a multi-modal fusion method with a policy routing mechanism to facilitate adaptive feature fusion. Finally, a classification head is employed to obtain the recognition results. Compared with existing mainstream multi-modal fusion algorithm frameworks, such as Transformer, our adoption of the computationally less complex heat conduction model achieves high accuracy while offering better computational efficiency and physical interpretability. Additionally, our newly proposed routing mechanism-guided multi-modal fusion strategy enables more effective integration of RGB-Event features.
The detailed implementation process will be described in the following sections.

\subsection{Network Architecture}

\subsubsection{Input Representation}   
\noindent
Given the RGB frames $\mathcal{I} \in \mathbb{R}^{B \times T \times C \times H \times W} = \{ I_1, I_2, ..., I_T \}$ and event streams $\mathcal{E}^{p} = \{ e_{1}, e_{2}, ..., e_{M} \}$, where $B$ and $T$ denote the batch size and the number of video frames, the dimensions of channel, height, and width are expressed as $C, H, W$, respectively, 
$e_j$ ($j \in [1, 2, ..., M]$) denotes each asynchronously launched event point, 
$M$ is the number of event points in the current sample. 
Each point $e_j$ exists in the form of a quadruple $\{x, y, t, p\}$, where $(x, y)$ denotes the spatial coordinates, $t$ and $p$ denote the time stamp and polarity. For the event streams $\mathcal{E}^{p}$, we stack them into event images based on the time stamp of the RGB frames, which can fuse more conveniently with the existing RGB modality. Consequently, we can obtain the multi-modal inputs with RGB frames $\mathcal{I} \in \mathbb{R}^{B \times T \times C \times H \times W}$ and aligned event frames $\mathcal{E} \in \mathbb{R}^{B \times T \times C \times H \times W}$.

\begin{figure*}
\center
\includegraphics[width=4.7in]{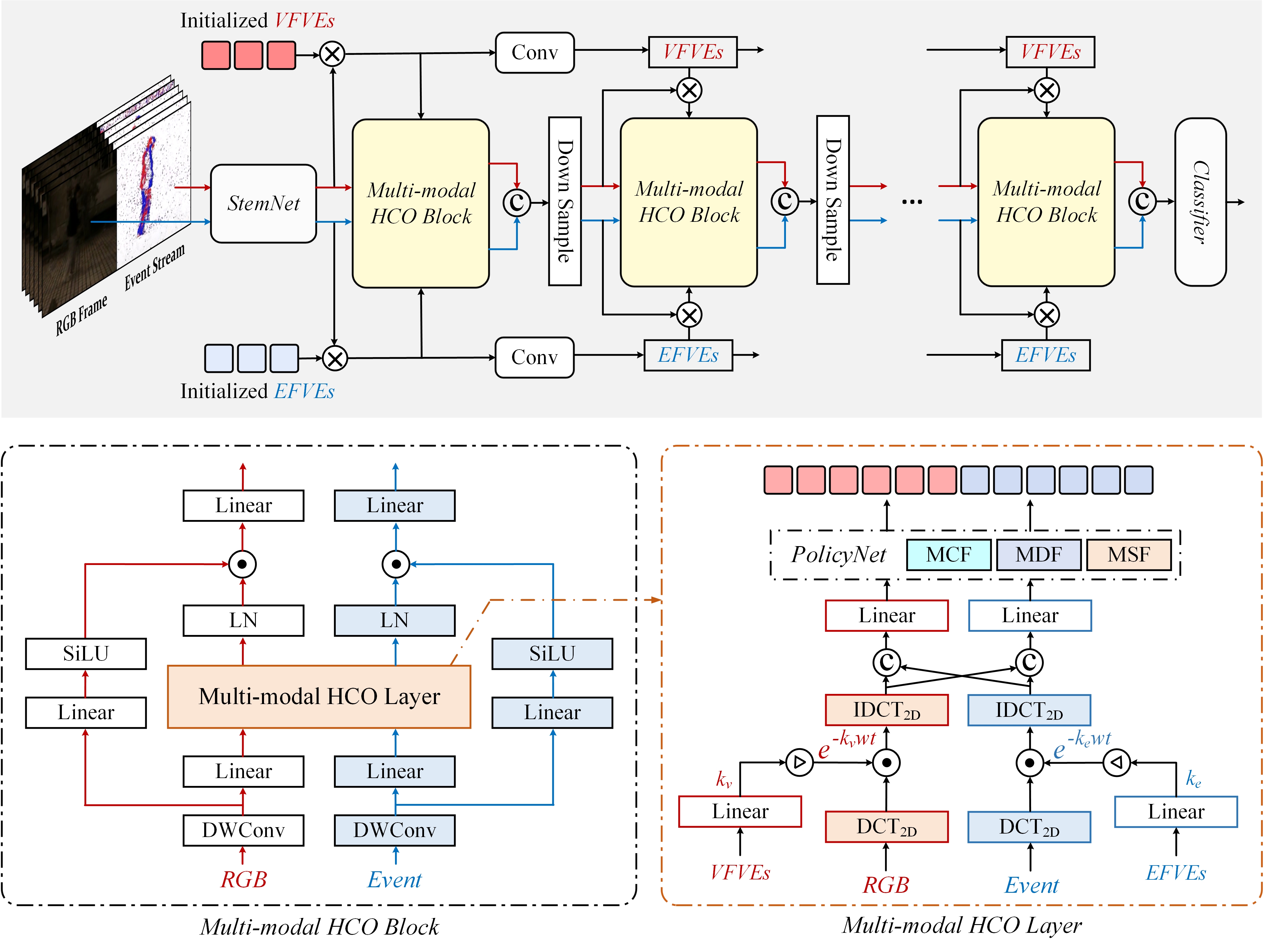}
\caption{An overview of our proposed MMHCO-HAR framework for RGB-Event based human activity recognition (HAR). A multi-modal visual heat conduction model is introduced to effectively integrate data from both RGB and event modalities to achieve robust HAR. Specifically, we propose modality-specific continuous Frequency Value Embeddings to capture the unique characteristics of each modality and enhance information interaction between multi-modal heat conduction blocks. Additionally, we introduce a policy routing based fusion method to adaptively fuse multi-modal information, ensuring optimized performance across diverse scenarios.
}\label{framework}
% \vspace{2in}
\end{figure*}

% \begin{equation}
% \begin{aligned}
% \label{Formula1}
% \frac{\partial u}{\partial t}=k\left(\frac{\partial^2u}{\partial x^2}+\frac{\partial^2u}{\partial y^2}\right),
% \end{aligned}
% \end{equation}
% by applying Fourier Transform to Formula~\ref{Formula1}, we can obtain:
% \begin{equation}
% \begin{aligned}
% \label{Formula2}
% \mathcal{F}\left(\frac{\partial u}{\partial t}\right)=k\mathcal{F}\left(\frac{\partial^2u}{\partial x^2}+\frac{\partial^2u}{\partial y^2}\right).
% \end{aligned}
% \end{equation}
% The form of $u (x, y, t)$ after Fourier Transform can be denoted $\widetilde{u}(\omega_{x},\omega_{y},t):=\mathcal{F}(u(x,y,t))$, and the left of Formula~\ref{Formula2} can be written as:
% \begin{equation}
% \label{Formula3}
%     \mathcal{F}\left(\frac{\partial u}{\partial t}\right)=\frac{\partial\widetilde{u}(\omega_x,\omega_y,t)}{\partial t}.
% \end{equation}

\subsubsection{Multi-modal Heat Conduction Backbone Network} 
As shown in Fig.~\ref{framework} (the top sub-figure), our RGB-Event HAR method is built on the vHeat model with physical interpretability. We begin by inputting both RGB frames $\mathcal{I} \in \mathbb{R}^{B \times T \times C \times H \times W}$ and event frames $\mathcal{E} \in \mathbb{R}^{B \times T \times C \times H \times W}$ to the StemNet, which consists of convolutional layers, BatchNorm layers, and GeLU activation functions, to obtain the image patches. Next, we feed the image patches into the multi-modal heat conduction blocks for feature fusion and interaction. As shown in Fig.~\ref{framework} (the bottom left sub-figure), the multi-modal HCO block contains two branches to deal with the bimodal input.

For each multi-modal HCO block, we first feed the RGB image patches and event image patches to the weight-sharing depth-wise convolutional networks. For RGB modality, a linear layer is employed to project the dimension of the feature channel from $C$ to $2*C$, and then divide the features into two parts according to the channel dimension. The first part is $X_R$, which is passed into a multi-modal HCO layer for thermal diffusivity based feature modeling (the detailed description will be introduced later), and $X_R'$ is obtained through a LayerNorm. The other part denoted as $Z_R$, is multiplied by $X_R'$ after passing through a linear layer and a SiLU activation function. Finally, we can obtain the output of this block through another linear layer. The event modality also performs the same operation as the RGB modality. Therefore, the formulas can be described as, 
\begin{equation}\begin{aligned}
\label{vheat_block_RGB}
 & X_R^{\prime\prime}=Linear(X_R^{\prime}*SiLU(Linear(Z_{R}))), \\
 & X_R^{\prime}=LN(MMHCO(X_{R})),
\end{aligned}\end{equation}

\begin{equation}\begin{aligned}
\label{vheat_block_Event}
 & X_E^{\prime\prime}=Linear(X_E^{\prime}*SiLU(Linear(Z_{E}))), \\
 & X_E^{\prime}=LN(MMHCO(X_{E})),
\end{aligned}\end{equation}
where $X_{E}$ denotes the event feature after the first linear layer and preparing to input a multi-modal HCO layer like $X_{R}$. The MMHCO block is visualized in the bottom right corner of Fig.~\ref{framework}.

After the processing of the depth-wise convolutional layer and linear layer, we can obtain the input $X_R$ of RGB modality and $X_{E}$ of event modality, respectively. Particularly, $DCT_{2D}$ and $IDCT_{2D}$ are the core operations of the MMHCO layer. 
The multi-modal features are passed through the $DCT_{2D}$ to output features in the frequency domain. Physically, the thermal diffusivity of different materials is different. Therefore, we design modality-specific Frequency Value Embeddings (FVEs) to predict the different thermal diffusivity $k$. Subsequently, the RGB and event features in the frequency domain will be multiplied by the coefficient decay matrix $e^{-k(\omega_x^2+\omega_y^2)t}$ respectively, and then the frequency domain features will be converted back to the time domain features through $IDCT_{2D}$. The formulas can be defined as follows,  
\begin{equation}
\label{DCT_IDCT_RGBE}
\begin{split}
    R_{out} &= \mathbf{IDCT_{2D}}(\mathbf{DCT_{2D}}(X_R) e^{-k(\omega_x^2 + \omega_y^2)t}), \\
    E_{out} &= \mathbf{IDCT_{2D}}(\mathbf{DCT_{2D}}(X_E) e^{-k(\omega_x^2 + \omega_y^2)t}).
\end{split}
\end{equation}
which are similar to the form of Formula~\ref{DCT_IDCT}. After that, we resort to the routing mechanism of the policy network to achieve adaptive feature fusion.

\subsubsection{Modality-specific Continuous FVEs}
The vHeat model introduced in \cite{wang2024vheat} is tailored to process a single modality input while emulating the phenomenon of heat conduction within a homogeneous material (unimodal). Within this architecture, the Frequency Value Embeddings (FVEs) are strategically initialized at each stage of the heat blocks, serving a role similar to that of position embeddings in visual Transformers without the frequency domain context. Given the challenge of handling multi-modal inputs, our approach must be adapted to account for the complex dynamics of heat transfer between two distinct materials, each representing a different modality, where their thermal diffusivity values differ significantly.

To bridge this gap, in this paper, modality-specific continuous FVEs are proposed to accomplish this objective. Our methodology begins with the independent random initialization of Visible Frequency Value Embeddings (VFVEs) and event Frequency Value Embeddings (EFVEs) for the two modalities. Subsequently, these embeddings are fused with the respective modal representations of RGB images and event streams. This fusion process yields tailored FVEs explicitly customized for each modality, thereby addressing the diverse properties inherent to different data types. Concurrently, a key observation we made was the arbitrary initialization of FVEs in the original vHeat design, which inadvertently led to a disconnect among blocks across stages due to their lack of interactivity. To address this limitation, we further design a mechanism to ensure that the FVEs in each subsequent stage are no longer reinitialized. Instead, they are propagated forward from the preceding stage via a convolutional projection layer. This innovative strategy not only ensures that each modality retains its individual thermal diffusivity $k$, reflecting its unique modal characteristics, but also significantly bolsters the interaction between FVEs across successive stages. By doing so, our enhanced multi-modal HCO model facilitates the cross-modal dynamical heat condution, ultimately leading to improved recognition performance in multi-modal settings.

\subsubsection{Policy Routing Strategy-guided Multi-modal Fusion} 

\begin{figure*}
\center
\includegraphics[width=4in]{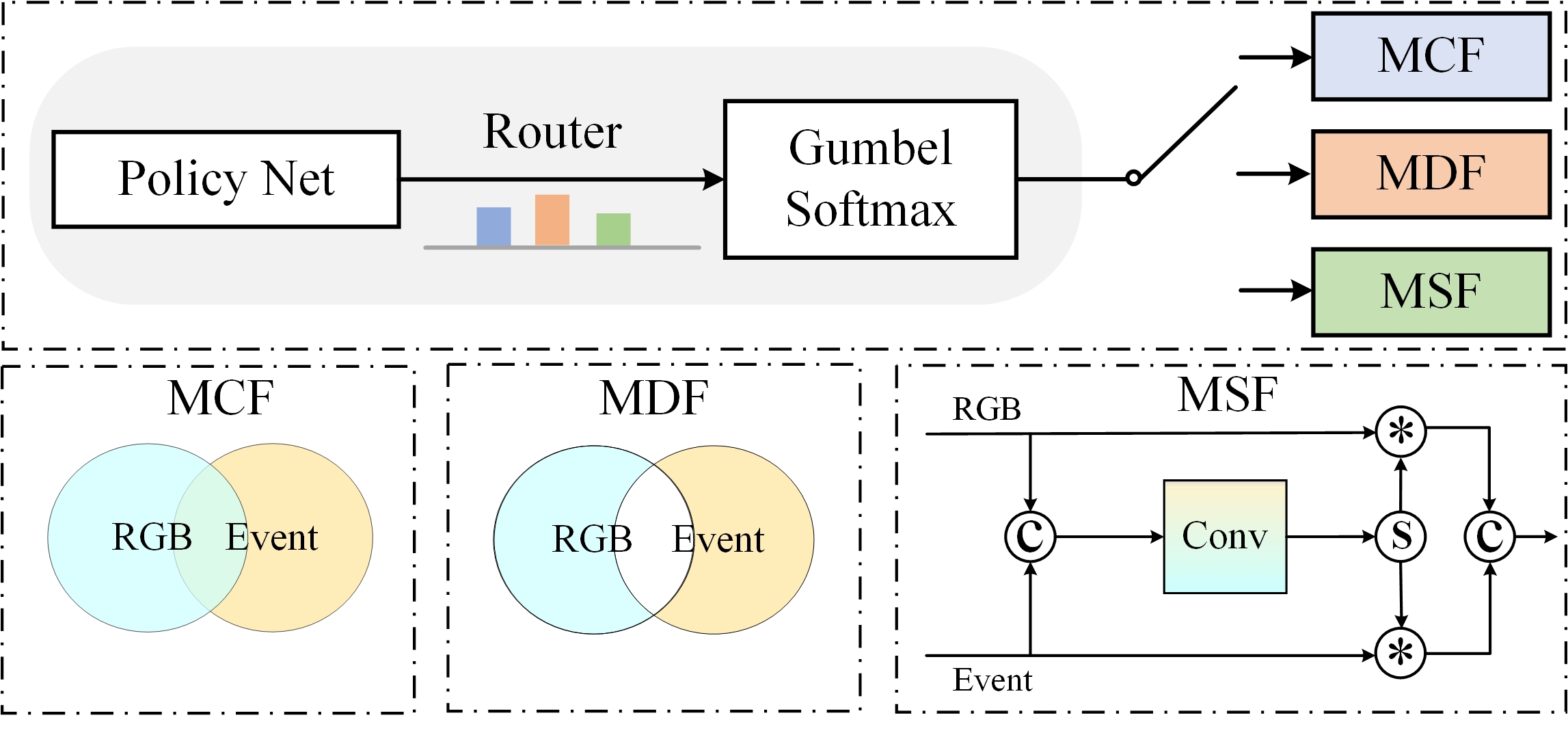}
\caption{The adaptive fusion module based on policy routing strategy, i.e., Modal Complementary Fusion (MCF), Modal Discriminative Fusion (MDF), and Modal Specific Fusion (MSF).}
\label{sub_framework}
\end{figure*}

Considering that different modalities have distinct characteristics, imbalanced multi-modal is an inevitable issue in multi-modal learning, which can result in suboptimal performance. Therefore, employing different fusion methods under various different conditions is reasonable. In this work, we propose three different fusion strategies to adapt to different situations. As shown in Fig.~\ref{sub_framework}, modal complementary fusion (MCF), modal discriminative fusion (MDF), and modal specific fusion (MSF) are considered. For RGB and event features ($F_R$ and $F_E$), a detailed explanation will be provided below.

\textbf{1). Modal Complementary Fusion (MCF)}: When both features exhibit strong individual performance, we employ the concatenation method to seamlessly combine them, leveraging the complementary strengths of each modality. This fusion strategy allows for the effective integration of spatial and temporal information. The formula for this fusion process is as follows,
\begin{equation}
\label{Fusion1}
    F_1 = Concat(F_R, F_E).
\end{equation}

\textbf{2). Modal Discriminative Fusion (MDF)}: Conversely, if the performance of either modality is suboptimal, we first identify the intersection between the two modalities through element-wise multiplication, which represents the common features shared by both. We then subtract this common feature from each modality, effectively eliminating low-quality shared features. This approach ensures that only the unique, high-quality features from each modality contribute to the final representation, minimizing the impact of poor performance in either modality. The formula is as follows, 

\begin{equation}
\label{Fusion2}
    F_2 = Concat\left( \left( F_R - F_R * F_E \right), \left( F_E - F_R * F_E \right) \right).
\end{equation}

\textbf{3). Modal Specific Fusion (MSF)}: In scenarios where one modality performs well while the other does not, we initially concatenate the two modalities to form a combined feature representation. Following this, we employ a simple convolutional network, which is paired with a sigmoid activation function, to learn and adaptively assign optimal weights to each modality based on their respective contributions. This approach enables a dynamic and weighted fusion of the modalities, allowing the network to emphasize the more informative modality while mitigating the impact of the less effective one, thereby improving overall performance. This process can be expressed as,
\begin{equation}
\label{Fusion3}
\begin{split}
    F_3 &= Concat(w_R * F_R, w_E * F_E), \\
    w_R, w_E &= \sigma(Conv(Concat(F_R, F_E))).
\end{split}
\end{equation}
where $w_R$ and $w_E$ denote the adaptively optimal weights, $\sigma$ is the $Sigmoid$ activation function.

Regarding the three optional fusion methods discussed above, previous approaches typically involve manually setting a fixed threshold to evaluate the quality of each modality and making a selection accordingly. In contrast, this work explores the use of a policy network for adaptive selection, thereby eliminating the need for introducing additional hyperparameters.
Concretely, our policy network is built using Multilayer Perceptrons (MLPs), which map the features extracted from the multi-modal HCO layer to $\mathcal{F} \in \mathbb{R}^{B \times 3 }$. Here, $B$ represents the batch size and $3$ indicates the number of possible selection paths. To maintain differentiability during the network’s backward process, we employ the Gumbel-Softmax~\cite{jang2016categorical} technique to obtain the one-hot vector representations. By optimizing the network, it is possible for the policy network to dynamically route a suitable path for multi-modal fusion, thereby achieving a better performance while eliminating the need for redundant manual configuration.

\subsection{Classification Head \& Loss Function}
Following the processing through multi-modal HCO blocks, we can achieve a robust multi-modal feature representation. Then, we input the fused features into the classification head to realize activity recognition. In particular, we initially acquire features $F_{out} \in \mathbb{R}^{B \times C }$ by applying a LayerNorm layer and average pooling layer. Subsequently, the features are mapped to the predicted scores $\hat{y_i} \in \mathbb{R}^{B \times C' }$ through a linear layer, where $C'$ is the number of action categories. Finally, we calculate the cross entropy loss between the predicted results and the ground truth labels, which can be formulated as,
\begin{equation}
\label{CE_loss} 
\mathcal{L} = -\frac{1}{N}\sum_{i=1}^N y_i\log\hat{y_i} + (1-y_i)\log(1-\hat{y_i}).
\end{equation}
where $\hat{y_i}$ is the predicted scores and ${y_i}$ is the ground truth.

\section{HARDVS 2.0 Benchmark Dataset}\label{sec4}

\subsection{Protocols}~\label{Protocols}
We aim to provide a good platform for the training and evaluation of multi-modal RGB-Event human activity recognition. When constructing the HARDVS 2.0 benchmark dataset, the following attributes/highlights are considered: 
\textbf{1). Large-scale:} As we all know, large-scale datasets play a very important role in the deep learning era. In this work, we collect more than 100k paired RGB-Event sequences to meet the needs for large-scale training and evaluation of HAR. 
\textbf{2). Wide varieties:} Thousands of human activities can exist in the real world, but existing DVS-based HAR datasets only contain limited categories. Therefore, it is hard to fully reflect the classification and recognition ability of HAR algorithms. Our newly proposed HARDVS 2.0 contains 300 classes that are several times larger than other datasets.  
\textbf{3). Different capture distances:} The HARDVS 2.0 dataset is collected under various distances, i.e., 1-2, 3-4, and more than 5 meters. 
\textbf{4). Long-term:} Most of the existing HAR datasets are microsecond-level, in contrast, each video in our HARDVS 2.0 dataset lasts for about 5 seconds.  
\textbf{5). Dual-modality:} The DAVIS346 camera can output both RGB frames and event streams, therefore, our dataset can be used for HAR by fusing video frames and events. In this work, we focus on HAR with both RGB and event modality and design a multi-modal method for HAR from a new perspective.

\begin{table*}
\center
\caption{Comparison of datasets for human activity recognition. MVW, MILL, MMO, DYB, OCC, and DR denote multi-view, multi-illumination, multi-motion, dynamic background, occlusion, and duration of the activities, respectively. Note that we only report these attributes of realistic DVS datasets for HAR..} \label{datasetlist} 
\resizebox{4.6in}{!}{
\begin{tabular}{l|ccccccccccccc}
\hline \toprule
\textbf{Dataset}    &\textbf{Year}  &\textbf{Scale}   &\textbf{Class}  &\textbf{Resolution}  &\textbf{Real}  &\textbf{MVW} &\textbf{MILL}  	&\textbf{MMO}  &\textbf{DYB} &\textbf{OCC} &\textbf{DR} \\ 
\hline 
\textbf{ASLAN-DVS}~\cite{lin1811temporal}   	 	&2011 	&$3,697$ 		&432 			&$240 \times 180$ 		&\xmark	&-    &- 		&- 		&-		&- 	&-  	\\ 
\textbf{MNIST-DVS}~\cite{serrano2015poker}   		 &2013   		 &$30,000$ 		&10 		&$128\times128$ 			&\xmark  			&-    &- 		&- 		&-		&- 	&-  	\\ 
\textbf{N-Caltech101}~\cite{orchard2015converting}  &2015  &$8,709 $		&101 		&$302 \times 245$ 			&\xmark 	 &-    &- 		&- 		&-		&- 	&-  \\ 
\textbf{N-MNIST}~\cite{orchard2015converting}   &2015  	&$70,000$ 		&10 		&$28\times28$ 			&\xmark  			&-    &- 		&- 		&-		&- 	&- 	\\ 
\textbf{CIFAR10-DVS}~\cite{li2017cifar10}   		 &2017 		&$10,000$ 		&10 		& $128\times128$			&\xmark  			&-    &- 		&- 		&-		&- 	&-  	\\ 
\textbf{HMDB-DVS}~\cite{kuehne2011hmdb}   	&2019  	 &$6,766$ 		&51 		&$240 \times 180$ 			&\xmark    &-    &- 		&- 		&-		&- 	&- 	\\ 
\textbf{UCF-DVS}~\cite{soomro2012dataset}   	&2019   &$13,320$  		&101 		&$240 \times 180$ 			&\xmark    &-    &- 		&- 		&-		&- 	&-  	\\ 
\textbf{N-ImageNet}~\cite{kim2021n} &2021 	&$1,781,167$   &1000   &$480 \times 640$   &\xmark  &-    &- 		&- 		&-		&- 	&- 	\\  
\textbf{ES-ImageNet}~\cite{lin2021imagenet}  &2021 	&$1,306,916$  &1000  		&$224 \times 224$  	 &\xmark  &-    &- 		&- 		&-		&- 	&- 	\\ 
\textbf{N-ROD}~\cite{cannici2021nrod} &2022 	&41,877   &51   &$640 \times 480$   &\xmark  &-    &- 		&- 		&-		&- 	&- 	\\ 
\hline 
\textbf{DvsGesture}~\cite{amir2017low}   	&2017 		 &$1,342$ 		&11 		&$128 \times 128$ 		&\cmark  			&\xmark    &\cmark 		&\xmark 		&\xmark			&\xmark  	& - 		\\ 
\textbf{N-CARS}~\cite{sironi2018hats} &2018  	&$24,029$ 		&2 		&$304 \times 240$  		&\cmark 	&\cmark    &\xmark 		&\cmark 		&\xmark			&\cmark 		 	&-  		\\ 
\textbf{ASL-DVS}~\cite{bi2020graph}	&2019 	&$100,800$ 		&24 			&$240 \times 180$ 		&\cmark  	&\xmark    &\xmark  &\xmark 		&\xmark 	&\xmark  	&0.1s  	\\ 
\textbf{PAF}~\cite{miao2019neuromorphic}	&2019 	 &$450$ 		&10 			&$346 \times 260$ 		&\cmark  	&\xmark    &\xmark  &\xmark 		&\xmark 	&\xmark  	&5s  	\\ 
\textbf{DailyAction}~\cite{liu2021event}  &2021 	 &$1,440$ 		&12 			&$346 \times 260$ 		&\cmark  	&\cmark    &\cmark  &\xmark 		&\xmark 	&\xmark  	&5s  	\\ 
\textbf{PokerEvent}~\cite{wang2023sstformer} &2024 		  &$27,102$ 		&114		&$346 \times 260$ 		&\cmark  &\cmark    &\cmark &\cmark &\cmark	&\cmark 	  	&-  	\\
\textbf{Dailydvs-200}~\cite{wang2024dailydvs} &2024  &$22,046$  &200	&$320 \times 240$  	&\cmark   &\cmark    &\cmark 		&\cmark 		&\cmark		&\cmark 	  	&1-20s  	\\ 
\hline 
\textbf{HARDVS 2.0} &2024 	&$107,646$ 		&300		&$346 \times 260$ 		&\cmark  &\cmark    &\cmark 		&\cmark 		&\cmark			&\cmark 	  	&5s  	\\ 
% \bottomrule
\hline \toprule
\end{tabular}
} 
\end{table*}

\begin{figure*} 
\center
\includegraphics[width=4.5in]{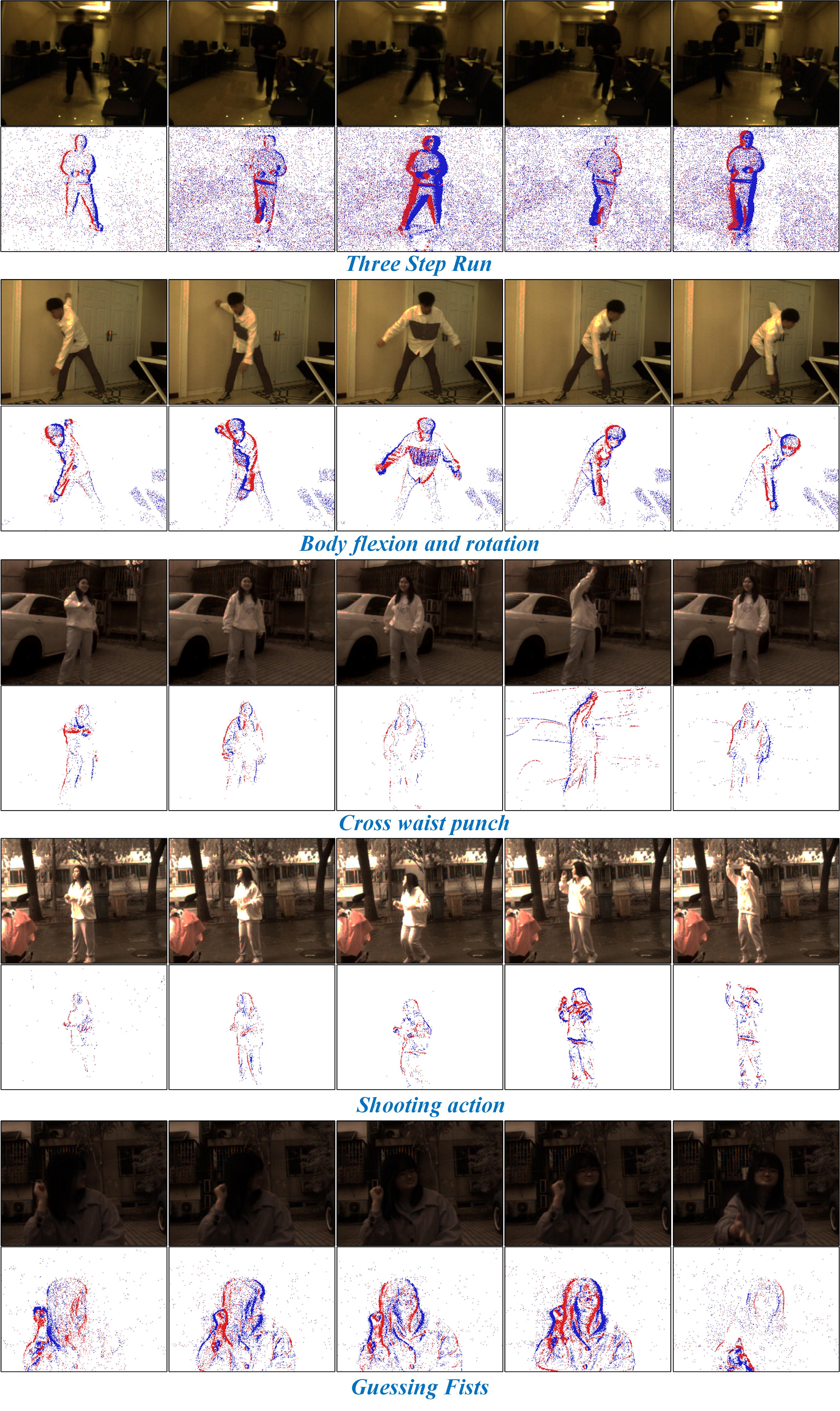}
\caption{Illustration of representative video clips in our HARDVS 2.0 dataset.}  
\label{HARDVS_samples2}
\end{figure*} 

Our dataset considers multiple challenging factors that may influence the results of HAR. The detailed introductions can be found below: 
\emph{(a). Multi-view:} We collect different views of the same behavior to mimic practical applications, including front-, side-, horizontal-, top-down-, and bottom-up-views. 
\emph{(b). Multi-illumination:} High dynamic range is one of the most important features of DVS sensors, therefore, we collect the videos under scenarios with strong-, middle-, and low-light. Note that, $60\%$ of each category are videos with low-light. Our dataset also contains many videos with \emph{glitter}, because we find that the DVS sensor is sensitive to flashing lights, especially at night. 
\emph{(c). Multi-motion:} We also highlight the features of DVS sensors by recording many activities with various motion speeds, such as slow-, moderate-, and high-speed. 
\emph{(d). Dynamic background:} As it is relatively easy to recognize activities without background objects, i.e., a stationary DVS camera, we also collect many activities with a dynamic moving camera to make our dataset challenging enough. 
\emph{(e). Occlusion:} In the real world, human activities can be occluded commonly and this challenge is also considered in our dataset.

\subsection{Data Collection and Statistic Analysis.} \label{dataCollection}
The HARDVS 2.0 dataset is collected with a DAVIS346 event camera whose resolution is $346 \times 260$. A total of five persons are involved in the data collection stage. From a statistical perspective, our dataset contains a total of $107,646$ paired video sequences and 300 classes of common human activities. We split $60\%, 10\%$, and $30\%$ of each category for training, validating, and testing, respectively. In total, the number of videos in the training, validating, and testing subsets are $64526$, $10734$, and $32386$, respectively. 
With the aforementioned characteristics, we believe our HARDVS 2.0 dataset will be a better evaluation platform for the neuromorphic classification problem, especially for the human activity recognition task. We carefully consider the aforementioned protocols when recording videos, ensuring that the unique characteristics of challenging scenarios are fully captured. 
A direct comparison with existing classification benchmark datasets can be found in Table~\ref{datasetlist} and Fig.~\ref{benchmarkCOMP}. 
We also give a visualization of partial categories and video clips of the HARDVS 2.0 dataset, as displayed in  Fig.~\ref{HARDVS_samples2}.

\section{Experiment}\label{sec5}

\subsection{Dataset and Evaluation Metric} 
\noindent
In this study, we evaluate our proposed model using two RGB-Event human activity recognition datasets: \textbf{PokerEvent}~\cite{wang2023sstformer}~\footnote{\url{https://github.com/Event-AHU/SSTFormer}} and our newly introduced \textbf{HARDVS 2.0}. 
The PokerEvent dataset is designed for recognizing character patterns on poker cards. It comprises 114 classes and contains 24,415 RGB-Event samples captured using a DVS346 event camera. The dataset is split into training and testing subsets, with 16,216 samples for training and 8,199 samples for testing.
We use the widely adopted evaluation metrics of \textbf{top-1} and \textbf{top-5 accuracy} for comparison with other methods.

\subsection{Implementation Details}
\noindent
Our proposed framework can be optimized in an end-to-end manner. The learning rate and weight decay are set as 0.001 and 0.0001, respectively. The SGD is selected as the optimizer and trained for a total of 30 epochs. In our implementations, a total of $24$ multi-modal HCO blocks are stacked as our backbone network like vHeat-B~\cite{wang2024vheat}. We redesign a new modality-specific continuous frequency value embedding and add an adaptive multi-modal fusion method based on the routing mechanism. Besides, we select 8 event frames as the input by following other benchmarked baselines.  
Our code is implemented using Python based on the PyTorch~\cite{paszke2019pytorch} framework, and the experiments are conducted on a server with CPU Intel(R) Xeon(R) Gold 5318Y CPU @2.10GHz and GPU RTX3090s. More details can be found in our source code.

\subsection{Comparison with Other SOTA Algorithms}

\noindent $\bullet$ \textbf{Results on HARDVS 2.0 Dataset.~}  
We first report the experimental results on the HARDVS 2.0 dataset. The comparison methods include CNN-baed (C3D~\cite{tran2015learning}, TSM~\cite{lin2019tsm}, X3D~\cite{feichtenhofer2020x3d}), Transformer-based (TimeSformer~\cite{bertasius2021space}, SSTFormer~\cite{wang2023sstformer}, ESTF~\cite{wang2024hardvs}), Large visual-language model-based (SAFE~\cite{li2023semantic}), and Mamba-based (Vision Mamba~\cite{zhu2024vision}). Instead, we explore a physical visual model based on heat conduction in this work. In addition to strong interpretability, we can see an extraordinary performance in the human activity recognition task, achieving 53.2\% on top-1 accuracy. Specifically, compared to the visual mamba model that has achieved great success in the field of vision, our improved visual heat conduction model surpasses it by +1.4\% on the top-1 accuracy. Furthermore, we also beat the Transformer-based methods like SSTFormer and ESTF thanks to the multi-modal HCO. Meanwhile, we also surpassed the SAFE method, which uses a Large visual-language model (LVLM) in the visual recognition tasks. Compared with the above outstanding methods, we can demonstrate the superiority of our proposed approach.

\begin{table}
\center
\small   
\caption{Results on the HARDVS 2.0 dataset (RGB+Event).} 
\label{HARDVS_acc}
% \resizebox{\columnwidth}{!}{ 
\begin{tabular}{l|l|l|l|c|c}
\hline \toprule 
\textbf{No.} & \textbf{Algorithm} & \textbf{Publish}  & \textbf{Backbone} &\textbf{Top-1} &\textbf{Top-5} \\
\hline
\#01 &\textbf{C3D}~\cite{tran2015learning}     &ICCV-2015   &3D-CNN  &50.9 &56.5    \\ 
\hline
\#02 &\textbf{TSM}~\cite{lin2019tsm}     &ICCV-2019   &ResNet-50  &52.6 & 62.1   \\ 
\hline
\#03 &\textbf{X3D}~\cite{feichtenhofer2020x3d}     &CVPR-2020   &ResNet  &47.4 &51.4    \\ 
\hline
\#04 &\textbf{TimeSformer}~\cite{bertasius2021space}    &ICML-2021   &ViT  &51.6  &58.5    \\ 
\hline
\#05 &\textbf{SSTFormer}~\cite{wang2023sstformer}     &arXiv-2023   &SNN-Former  &53.0  &60.1    \\ 
\hline 
% \#06 &\textbf{TSCFormer}~\cite{wang2023unleashing}    &arXiv-2023   &CNN-Former  &53.0 &\textbf{62.7}    \\ 
% \hline
\#06 &\textbf{SAFE}~\cite{li2023semantic}     &PR-2024   &VLM  &50.1 &56.1    \\ 
\hline
\#07 &\textbf{ESTF}~\cite{wang2024hardvs}     &AAAI-2024   &ResNet-Former  &49.9 &55.8    \\ 
\hline
\#08 &\textbf{Vision Mamba}~\cite{zhu2024vision}     &ICML-2024   &Mamba  &51.8 &59.5    \\ 
\hline
\#09 &\textbf{Ours}     &-   &Multi-modal HCO  & 53.2   & 62.1    \\ 
\hline \toprule 
\end{tabular}
% }
\end{table}

\noindent $\bullet$ \textbf{Results on PokerEvent~\cite{wang2023sstformer} Dataset.~}  
PokerEvent is a special recognition dataset that records the poker cards' character patterns by a DAVIS346 camera. As shown in Table~\ref{PokerEvent_acc}, we compare the experimental results on the PokerEvent dataset with other methods. Obviously, our method surpasses previous mainstream algorithms like TSM~\cite{lin2019tsm}, TAM~\cite{liu2021tam}, SSTFormer~\cite{wang2023sstformer}, to a certain extent, achieving 57.4\% on the top-1 accuracy. Consequently, the prominent performance is achieved on the PokerEvent dataset due to the modality-specific continuous FVEs and policy routing based fusion method. This implies that the heat conduction multi-modal model based on physical prior is effective for conventional visual recognition tasks. 

\begin{table}
\center
\small   
\caption{Results on the PokerEvent dataset (RGB+Event).} 
\label{PokerEvent_acc}
% \resizebox{\columnwidth}{!}{ 
\begin{tabular}{l|l|c|c|c}
\hline \toprule 
\textbf{No.} & \textbf{Algorithm}   & \textbf{Publish}  & \textbf{Backbone}  & \textbf{Top-1}  \\
\hline
\#01 &\textbf{C3D}~\cite{tran2015learning}     &ICCV-2015   &3D-CNN  &51.8    \\ 
\hline
\#02 &\textbf{TSM}~\cite{lin2019tsm}     &ICCV-2019   &ResNet-50  &55.4    \\ 
\hline
\#03 &\textbf{ACTION-Net}~\cite{wang2021action}     &CVPR-2021   &ResNet-50  &54.3 \\ 
\hline
\#04 &\textbf{TAM}~\cite{liu2021tam}     &ICCV-2021   &ResNet-50  &53.7    \\ 
\hline
\#05 &\textbf{V-SwinTrans}~\cite{liu2022video}  &CVPR-2022   &Swin Transformer  &54.1 \\ 
\hline
\#06 &\textbf{TimeSformer}~\cite{bertasius2021space}     &ICML-2021   &ViT  &55.7   \\ 
\hline
\#07 &\textbf{X3D}~\cite{feichtenhofer2020x3d}     &CVPR-2020   &ResNet  &51.8    \\ 
\hline
\#08 &\textbf{MVIT}~\cite{li2022mvitv2}     &CVPR-2022   &ViT  &55.0    \\ 
\hline
\#09 &\textbf{SSTFormer}~\cite{wang2023sstformer}     &arXiv-2023   &SNN-Former  &54.7    \\ 
\hline
% \#10 &\textbf{TSCFormer}~\cite{wang2023unleashing}      &arXiv-2023   &CNN-Former  &\textbf{57.7}   \\ 
% \hline
\#10 &\textbf{SAFE}~\cite{li2023semantic}      &PR-2024   &VLM  &57.6  \\
\hline
\#11 &\textbf{Ours}      &-   &Multi-modal HCO  &57.4 \\
\hline \toprule 
\end{tabular}
% }
\end{table}

\subsection{Component Analysis} 
\noindent
In this section, we will isolate each component to analyze our MMHCO-HAR framework on our HARDVS 2.0 dataset. As shown in Table~\ref{CAResults}, the vHeat denotes only employing the vHeat model to replace the original backbone network and fusing bimodal features through the concatenate operation, achieving an accuracy of 52.3\% on acc/top-1. To enhance modality-specific visual heat conduction and improve information interaction between blocks, we design modality-specific continuous FVEs, which improves the results to 52.8\% acc/top-1. Concurrently, we propose three fusion strategies for different feature combinations and introduce policy networks to adaptively routing and select the appropriate fusion method. In this way, there has been a certain improvement compared to the initial version, and the issue of imbalance in multi-modal has been alleviated. Finally, combining the above innovations, we develop a multi-modal HCO model, which achieves even higher accuracy on both acc/top-1 and acc/top-5 metrics on the HARDVS 2.0 dataset. Through experimental analysis of the above components, it is evident that our proposed modality-specific continuous FVEs and policy routing adaptive fusion mechanism significantly aid in learning multi-modal human activities, demonstrating the effectiveness of our method for HAR tasks.

\subsection{Ablation Study}  

\begin{figure*} 
\center
\includegraphics[width=5in]{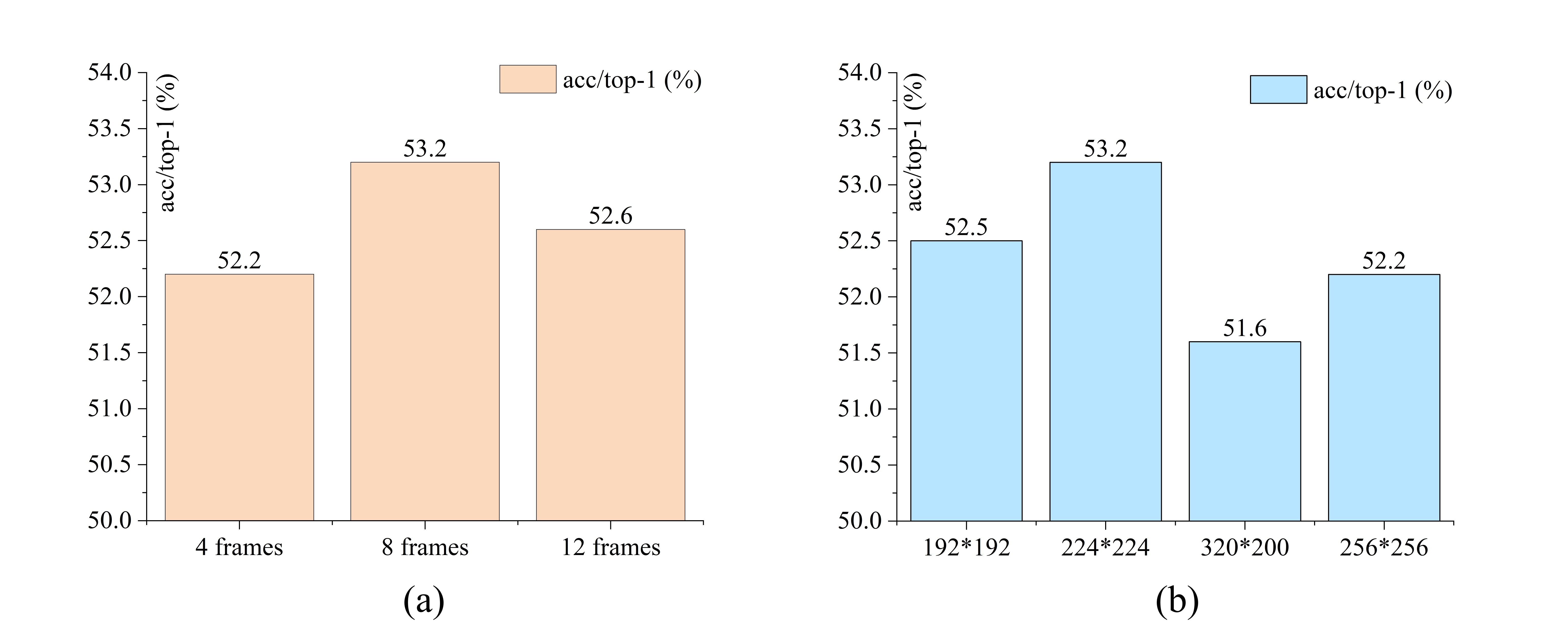}
\caption{(a). The impact of the number of frames on accuracy; 
(b). The impact of input image resolution on accuracy. }  
\label{ablation_study_imgs}
\end{figure*}

\noindent $\bullet$ \textbf{Analysis of Number of Input Frames.~} 
In this section, we analyze the influence of various experimental parameters through ablation studies. Firstly, we examine the effect of the number of input frames on model performance. As illustrated in Fig.~\ref{ablation_study_imgs} (a), we conduct experiments using 4, 8, and 12 input frames per modality for comparison. The results indicate that the best recognition accuracy is achieved when each modality inputs 8 frames. We believe that using fewer frames limits the model’s ability to capture sufficient temporal information, leading to inaccurate video recognition. Conversely, inputting too many frames increases the risk of overfitting, as the model may focus on redundant or noisy features.

\begin{table} 
\center
% \small   
\caption{Component Analysis on our HARDVS 2.0 Dataset. MSC denotes modality-specific continuous FVEs; PRF denotes policy routing based fusion method} \label{CAResults} 
% \resizebox{0.9\textwidth}{!}{ 
\begin{tabular}{ccc|cc} 		%% \xmark   \cmark  
\hline \toprule 
\textbf{vHeat}  &\textbf{MSC} &\textbf{PRF} &\textbf{acc/top-1} &\textbf{acc/top-5} \\
\hline 
\cmark   &          &               &52.3                  &60.2          \\
\cmark   &\cmark    &               &52.8                  &61.1         \\
\cmark   &          &\cmark         &52.7                  &60.7            \\
\cmark   &\cmark    &\cmark         &\textbf{53.2}         &\textbf{62.1}         \\
\hline \toprule  
\end{tabular}
% }
\end{table}

\noindent $\bullet$ \textbf{Analysis of Input Resolution.~}
Next, we analyze the effect of input image resolution on the model's recognition accuracy, as illustrated in Fig.~\ref{ablation_study_imgs} (b). The results show that the model achieves optimal performance when the input resolution is set to $224 \times 224$. Interestingly, increasing the resolution to $256 \times 256$ not only introduces a heavier computational burden but also leads to a decline in recognition accuracy, which is counterintuitive. We attribute this performance drop to the fixed receptive field of the network. As the input resolution increases, the receptive field covers a smaller proportion of the image, limiting the network's ability to effectively capture and predict foreground targets across different scales. This scale mismatch hinders the model’s capacity to extract meaningful features, ultimately resulting in reduced accuracy. To address this issue, we align the input image resolution with the network's receptive field size throughout this study, ensuring balanced feature extraction and optimal recognition performance.

\begin{table}
\centering
\small
\caption{Ablation Studies of Input modalities on HARDVS 2.0 dataset.}
\label{ablation_study1}
\resizebox{0.85\textwidth}{!}{ 
\begin{tabular}{l|c|c}
\hline \toprule 
\textbf{\# Input modalities} & \textbf{acc/top-1}  & \textbf{acc/top-5}  \\
\hline
1. RGB frame only                   & 49.2  & 53.8 \\
2. Event streams only                & 51.2   & 57.5 \\
3. \textbf{Both RGB frame and Event streams}     & \textbf{53.2}   & \textbf{62.1} \\
\hline \toprule 
\end{tabular}
}
\end{table}

\noindent $\bullet$ \textbf{Analysis of the Input modalities.~}
As shown in Table~\ref{ablation_study1}, we further analyze the impact of different input modalities on human activity recognition tasks. Interestingly, when using only event data as input, the recognition accuracy surpasses that of using only RGB data. This result suggests that our HARDVS 2.0 dataset effectively capitalizes on the strengths of event-based data, demonstrating that in many scenarios, event data can outperform traditional RGB data. This finding underscores the advantages of event cameras, particularly in capturing dynamic scenes or handling challenging lighting conditions. Moreover, when both RGB and event data are combined as input, the recognition accuracy improves even further, achieving a 2\% increase in top-1 accuracy compared to using only unimodal event data. This highlights the complementary nature of RGB and event modalities and the effectiveness of multi-modal fusion in enhancing human activity recognition performance.

\begin{table}[h]
\centering
\small
\caption{Ablation Studies of Fusion Methods on HARDVS 2.0 dataset.}
\label{ablation_study2}
\resizebox{0.85\textwidth}{!}{ 
\begin{tabular}{l|c|c}
\hline \toprule 
\textbf{\# Fusion Methods} & \textbf{acc/top-1} & \textbf{acc/top-5} \\
\hline
1. Simple addition      & 52.3  & 59.3  \\
2. Concatenate          & 52.3 & 60.2  \\
3. Random selection     & 52.6  & 60.3   \\
4. \textbf{Adaptive routing selection}   & \textbf{53.2}   & \textbf{62.1}   \\
\hline \toprule 
\end{tabular}
}
\end{table}

\noindent $\bullet$ \textbf{Analysis of Fusion Methods.~} 
In this study, we propose a novel policy routing based fusion method to effectively tackle the challenge of imbalanced multi-modal. As illustrated in Table~\ref{ablation_study2}, we first explored basic fusion strategies such as simple addition and concatenation. However, both methods achieved a top-1 accuracy of 52.3\%, highlighting their limitations in capturing complex multi-modal interactions. To enhance adaptability across diverse scenarios, we designed three distinct fusion strategies tailored to better integrate complementary modality information. Notably, the results presented in the third and fourth rows of Table~\ref{ablation_study2} demonstrate that our policy routing based approach significantly outperforms the strategy of randomly selecting among the three fusion methods, showcasing superior robustness and generalization capabilities. These compelling experimental findings validate the effectiveness of our proposed method in advancing performance on multi-modal human activity recognition tasks.

\noindent $\bullet$ \textbf{Efficiency Analysis.~} 
As shown in Table~\ref{Efficiency_Analysis}, we conduct a comprehensive comparison of three critical model metrics: checkpoint storage size, computational efficiency (measured in FLOPs), and parameter count. The results demonstrate that our proposed multi-modal HCO achieves a reduction in model storage requirements (from 1.04GB to 309.3MB) compared to the multi-modal Transformer (Here, we employ a 10-layer stacked Transformer architecture for comparative analysis), improves computational efficiency by 1.4× (from 88.3GFLOPs to 61.7GFLOPs), and maintains superior performance while using only 55\% of the parameters (76.1M vs 138.5M). This significant advantage stems from our novel heat conduction-based multi-modal model.

\begin{table}[h]
\centering
\small
\caption{Comparison of Model Storage Size, Computational Efficiency, and Parameters.}
\label{Efficiency_Analysis}
\resizebox{0.85\textwidth}{!}{ 
\begin{tabular}{l|c|c|c}
\hline \toprule 
\textbf{\# Models} & \textbf{Storage Size} & \textbf{FLOPs} & \textbf{Params}\\
\hline
Multi-modal HCO     & 309.3MB  & 61.7G & 76.1M \\
Multi-modal Former          & 1.04GB & 88.3G & 138.5M \\

\hline \toprule 
\end{tabular}
}
\end{table}

\subsection{Visualization}

\begin{figure*} 
\center
\includegraphics[width=4.6in]{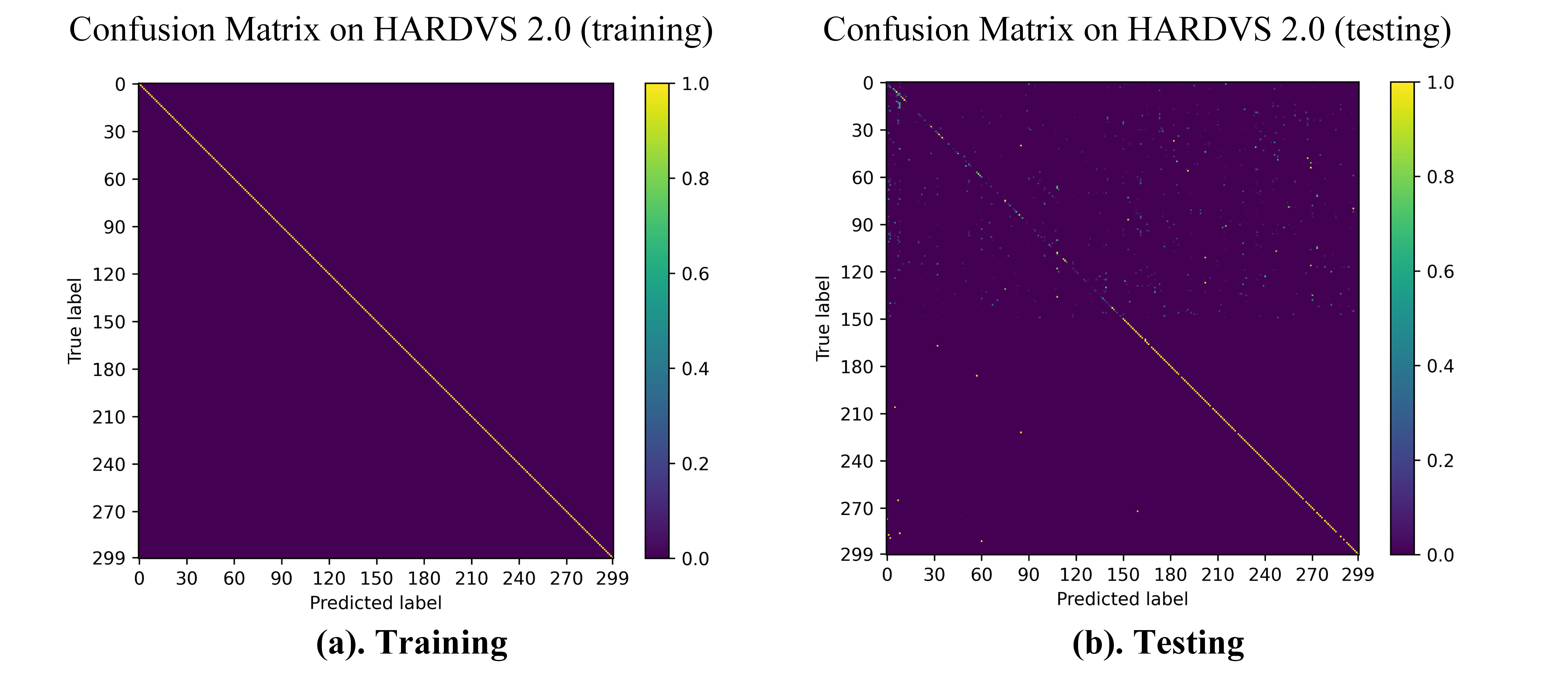}
    \caption{Visualization of confusion matrix on our proposed HARDVS 2.0 dataset.}  
\label{confusion_matrix}
\end{figure*}

\noindent $\bullet$ \textbf{Confusion Matrix on HARDVS 2.0.~}
In this section, we will provide some visualizations to help readers better understand our dataset and method. As shown in Fig.~\ref{confusion_matrix} (a) and Fig.~\ref{confusion_matrix} (b), we present the confusion matrix on the training and testing subsets of our HARDVS 2.0 dataset. It can be seen that our model performs much better on the training set than on the testing set, which indicates that our dataset is highly challenging and there is room for further optimization of our method on the test set.

\begin{figure*} 
\center
\includegraphics[width=4.8in]{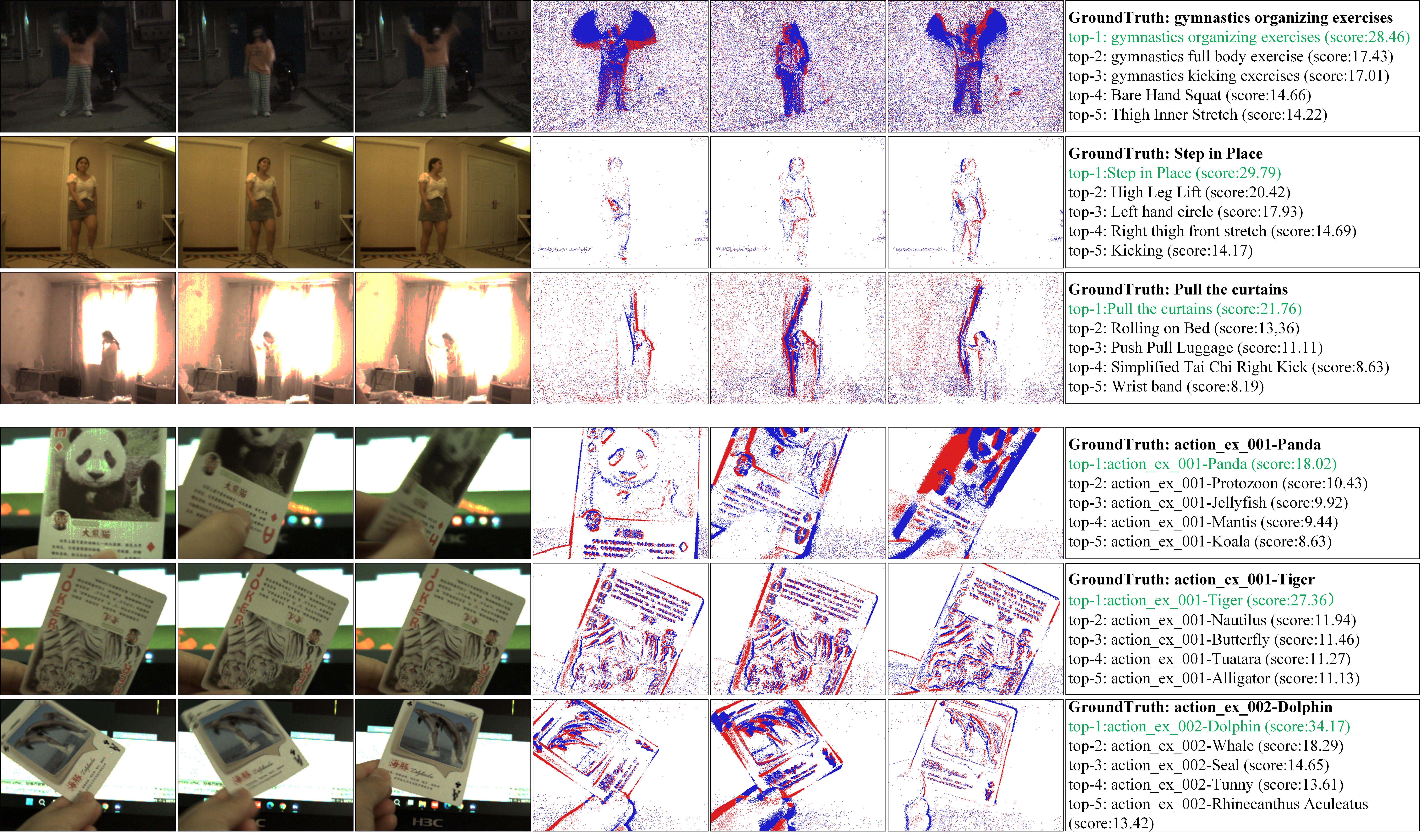}
    \caption{Visualization of Top-5 accuracy on the HARDVS 2.0 and the PokerEvent dataset.}  
\label{top5_acc}
\end{figure*}

\noindent $\bullet$ \textbf{Top-5 Predictions on HARDVS 2.0 and PokerEvent.~}
As shown in Fig.~\ref{top5_acc}, we also present the top-5 prediction results on the HARDVS 2.0 and PokerEvent datasets. For visualization, three categories are selected from each dataset, including \emph{set of broadcast gymnastics organizing exercises}, \emph{Step in Place}, \emph{Pull the curtains}, \emph{Panda}, \emph{Tiger}, and \emph{Dolphin}. The prediction results demonstrate that our method can accurately identify categories that match the ground truth, highlighting its effectiveness in multi-modal human activity recognition.

\begin{figure*} 
\center
\includegraphics[width=4.5in]{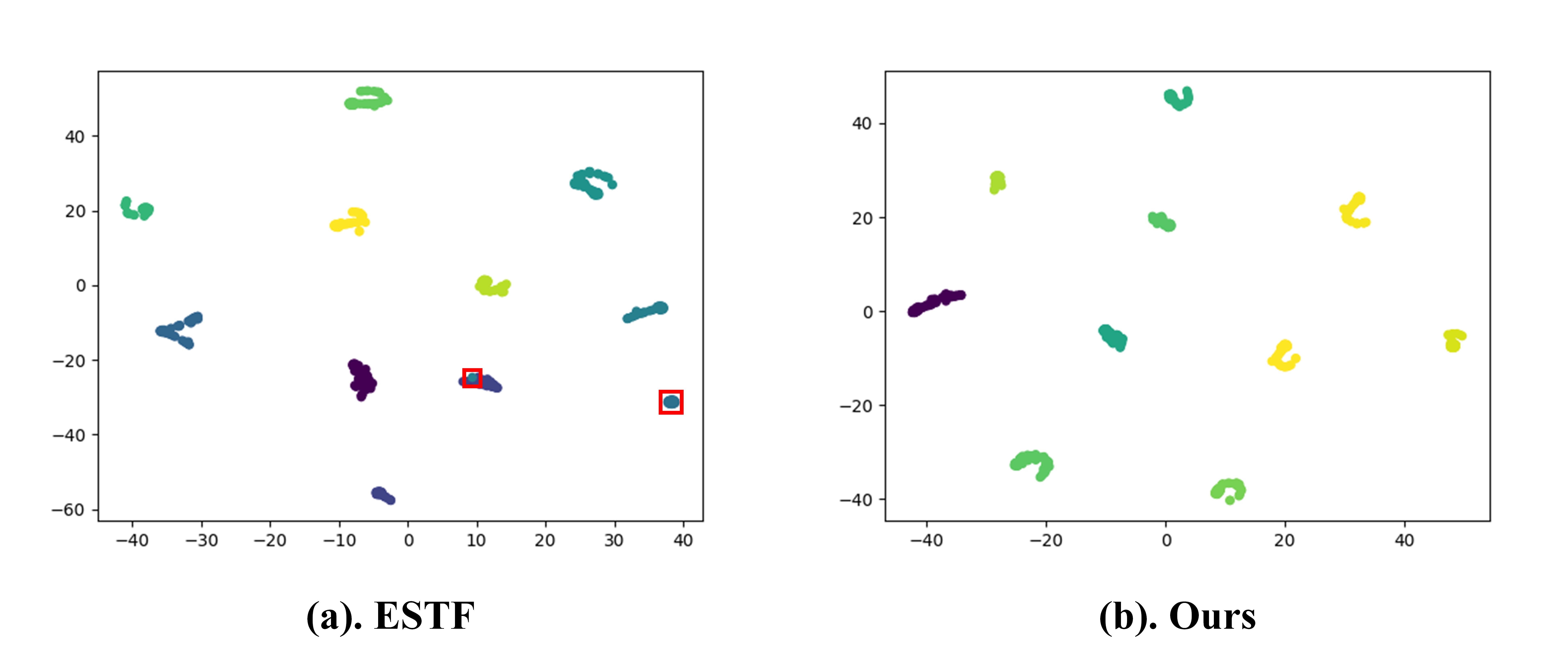}
\caption{Visualization of feature distribution of multi-modal ESTF and Ours.}  
\label{feature_distribution}
\end{figure*}

\noindent $\bullet$ \textbf{Feature Distribution of ESTF and Ours.~}
As depicted in Fig.~\ref{feature_distribution} (a) and (b), we illustrate the feature distribution of multi-modal ESTF and our proposed method. ESTF is constructed using a spatiotemporal Transformer, whereas our approach is a novel multi-modal visual model based on heat conduction. We randomly select 10 categories from the HARDVS 2.0 dataset, and the feature clustering performance of our method is clearly superior to that of ESTF. This result further highlights the effectiveness and robustness of our proposed multi-modal HCO model.

\begin{figure*} 
\center
\includegraphics[width=4.6in]{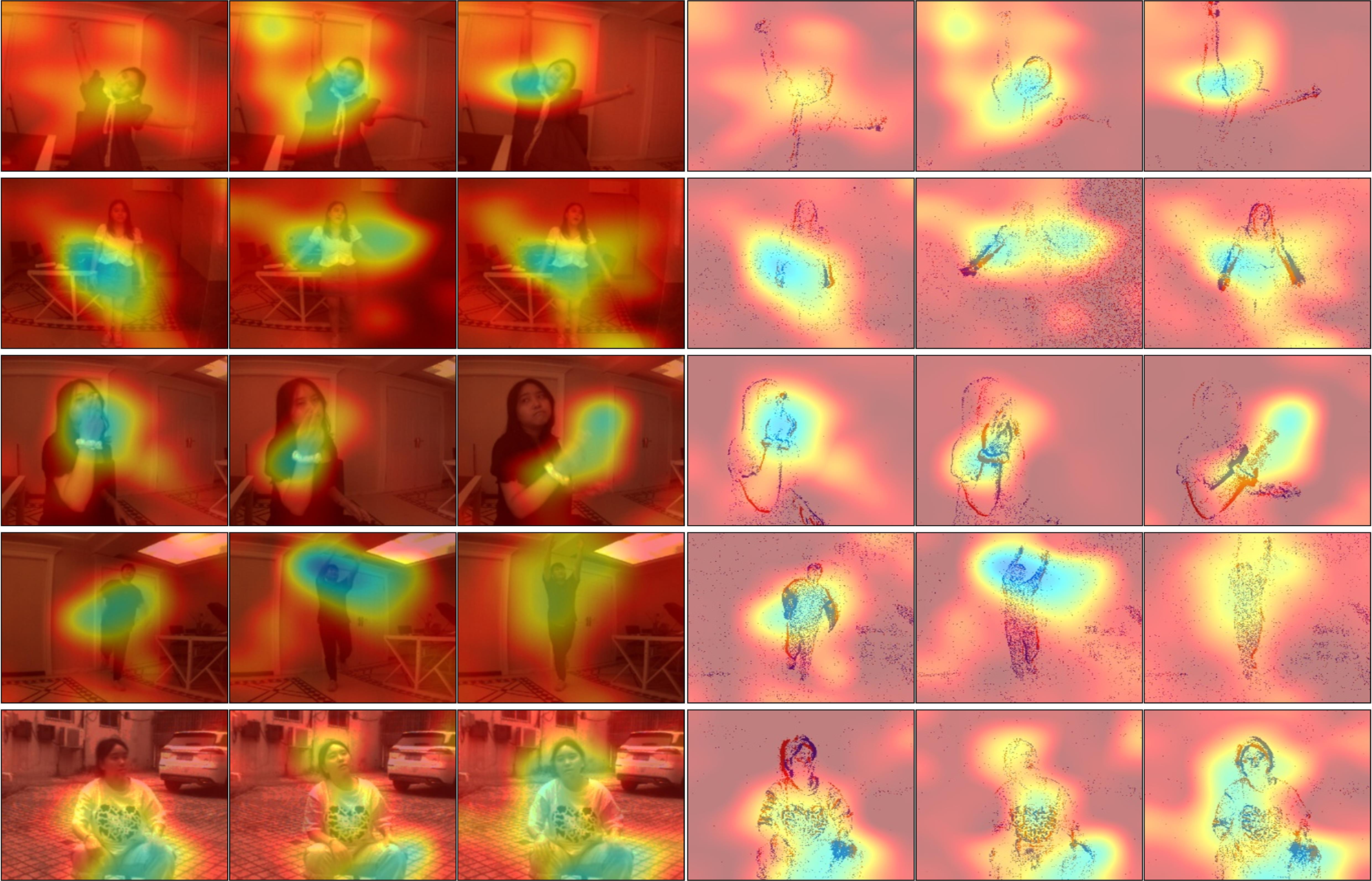}
    \caption{Activation maps on our proposed HARDVS 2.0 dataset.}  
\label{feature_map}
\end{figure*}

\noindent $\bullet$ \textbf{Activation Maps on HARDVS 2.0 Dataset.~}
As presented in Fig.~\ref{feature_map}, we provide the heatmaps of the algorithm's activation to demonstrate the attention effect of our method on human activities, where the dark blue regions highlight the areas of the image that the algorithm focuses on the most. It is evident that on the proposed HARDVS 2.0 dataset, our method can accurately capture human activities in videos, showcasing the robustness of our method in both locating and recognizing human activities.

\subsection{Limitation Analysis}  
\noindent
While this study introduces a novel multi-modal visual heat conduction model as an efficient alternative to Transformer-based architectures for improved human activity recognition, our current implementation does not fully exploit the inherent advantages of event cameras (particularly their low latency and high dynamic range) to optimize cross-modal fusion.
% we observed no significant improvement in efficiency compared to Transformer networks (the original paper points out that the complexity of the HCO is O(\(N^{1.5}\))). 
Furthermore, this work does not leverage additional textual semantic information or integrate large language models to guide the multi-modal visual learning process. The incorporation of such techniques could potentially enhance the model’s ability to better understand and align the visual and textual modalities. In our future work, we aim to address these limitations by further optimizing the multi-modal HCO model through integrating the advantages of event cameras, and exploring the synergy between textual information and visual modalities, with the goal of improving both efficiency and performance in multi-modal human activity recognition tasks.

\section{Conclusion and Future Works}~\label{sec6}
\noindent
In this paper, a novel RGB-Event based multi-modal human activity recognition framework has been proposed, termed MMHCO-HAR. We begin by exploring a heat conduction-based visual model and extend it to the multi-modal heat conduction version. Specifically, modality-specific continuous Frequency Value Embeddings are introduced to better accommodate the unique characteristics of different modalities and enhance information interaction between multi-modal heat conduction blocks. Additionally, we also propose a policy routing mechanism for adaptive multi-modal fusion, which helps address the issue of imbalanced multi-modal. Furthermore, we extend the original dataset into a bimodal version and introduce HARDVS 2.0, a large-scale multi-modal HAR dataset, to fill the data gap in this research field. HARDVS 2.0 contains 300 categories and includes 107,646 paired RGB-Event video samples, covering a range of challenging attributes such as multi-view, low-light, fast-motion, and so on. We hope that our methods and dataset will drive progress in this area and contribute to the research community. 

In our future works, we will explore how to leverage the low latency characteristics of event cameras to achieve more efficient activity recognition. Additionally, we intend to employ multi-modal large models to integrate text and other modalities, enhancing semantic understanding.
% In future work, we will continue to explore new ideas and approaches for multi-modal HAR, striving for even more accurate and efficient activity recognition.

%%===========================================================================================%%
%% If you are submitting to one of the Nature Portfolio journals, using the eJP submission   %%
%% system, please include the references within the manuscript file itself. You may do this  %%
%% by copying the reference list from your .bbl file, paste it into the main manuscript .tex %%
%% file, and delete the associated \verb+\bibliography+ commands.                            %%
%%===========================================================================================%%

\bibliography{reference}% common bib file
%% if required, the content of .bbl file can be included here once bbl is generated
%%\input sn-article.bbl

%% Default %%
%%\input sn-sample-bib.tex%

\end{document}